\documentclass[letterpaper, 10 pt, conference]{IEEEconf}  

\IEEEoverridecommandlockouts                              

\usepackage{graphics} 
\usepackage{epsfig} 
\usepackage{times} 
\usepackage{amsmath} 
\usepackage{amssymb}  
\usepackage{amsfonts}
\usepackage{subfigure}
\usepackage[ruled,vlined]{algorithm2e}
\usepackage{caption}
\usepackage{cite}
\usepackage{graphics} 
\usepackage{epsfig} 
\usepackage{times} 
\usepackage{amsmath} 
\usepackage{amssymb}  
\usepackage{amsfonts}
\usepackage{subfigure}
\usepackage{caption}
\usepackage{cite}
\usepackage{threeparttable}
\usepackage{booktabs}
\usepackage{makecell}

\title{\LARGE \bf
Leveraging Extrinsic Dexterity for Occluded Grasping 
\\ on Grasp Constraining Walls
}

\author{Keita Kobashi$^{1}$ and Masayoshi Tomizuka$^{1}$
\thanks{$^{1}$Keita Kobashi and Masayoshi Tomizuka are in Department of Mechanical Engineering,
        University of California, Berkeley, CA, 94704
}
}

\begin{document}

\maketitle
\thispagestyle{empty}
\pagestyle{empty}

\begin{abstract}
This study addresses the problem of occluded grasping, where primary grasp configurations of an object are not available due to occlusion with environment.  
Simple parallel grippers often struggle with such tasks due to limited dexterity and actuation constraints.
Prior works have explored object pose reorientation such as pivoting by utilizing extrinsic contacts between an object and an environment feature like a wall, to make the object graspable.
However, such works often assume the presence of a short wall, and this assumption may not always hold in real-world scenarios. 
If the wall available for interaction is too large or too tall, the robot may still fail to grasp the object even after pivoting, and the robot must combine different types of actions to grasp.
To address this, we propose a hierarchical reinforcement learning (RL) framework.
We use Q-learning to train a high-level policy that selects the type of action expected to yield the highest reward.
The selected low-level skill then samples a specific robot action in continuous space.
To guide the robot to an appropriate location for executing the selected action, we adopt a Conditional Variational Autoencoder (CVAE).
We condition the CVAE on the object point cloud and the skill ID, enabling it to infer a suitable location based on the object geometry and the selected skill.
To promote generalization, we apply domain randomization during the training of low-level skills.
The RL policy is trained entirely in simulation with a box-like object and deployed to six objects in real world.
We conduct experiments to evaluate our method and demonstrate both its generalizability and robust sim-to-real transfer performance with promising success rates.

\end{abstract}

\section{INTRODUCTION}
\begin{figure}[t]
 \centering
    \includegraphics[clip, width=0.99\columnwidth]{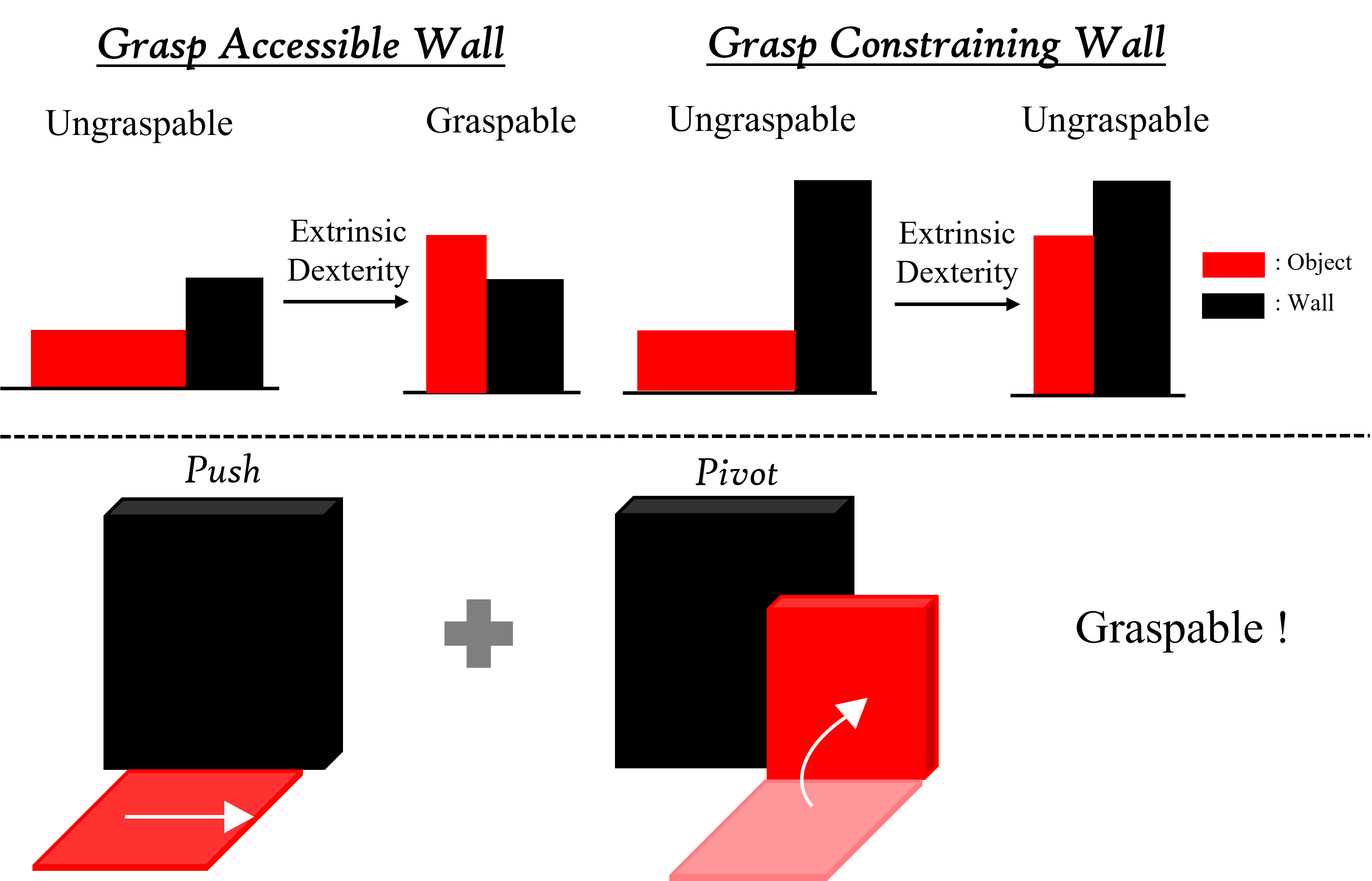}
 \caption{The problem description. In this work, we address an occluded grasping problem on grasp constraining walls. Previous works implicitly assume grasp accessible walls, where the robot can grasp an object after pivoting. However, if the wall is too large or too tall, the robot cannot grasp even after pivoting. To deal with this problem, we consider combining pivoting actions and pushing actions to make the object graspable even on grasp constraining walls.}
 \label{fig: problem_example}
\end{figure}

Robotic grasping is one of the fundamental tasks and is important to address.
Typically, when the robot picks and places an object in another location or the robot performs further manipulation, such as robotic insertion, the robot needs to grasp the object \cite{zhang2022learning, zhang2023efficient, fuchioka2024robotic, inoue2017deep, wang2019manipulation, liang2024robust}.
Whether the object can be grasped or not can depend on the object pose.
For instance, when grasping a flat box on a table (as shown in Fig. \ref{fig: problem_example}), the robot may fail due to the gripper's actuation limits.
In this work, we address such an occluded grasping problem \cite{zhou2023learning, wang2024multi}, which deals with grasping an object whose primary grasp configurations are occluded.

A common approach to handle occlusion is to exploit extrinsic contacts between an object and an environment to reorient the object into a graspable pose.
Although some prior works address similar problems \cite{zhang2023efficient, zhang2023learning, zhou2023learning, yang2024learning}, they implicitly assume a grasp accessible environment, where the robot can grasp the object after pivoting motion as shown in Fig. \ref{fig: problem_example}.
In this work, we consider a grasp constraining environment; the grasping configurations are still occluded even after pivoting the object.
In such an environment, the robot needs to combine different types of actions in addition to pivoting to grasp the object.

The main difficulties of this problem lie in 1) handling complex physical interactions, 2) switching different types of actions automatically, 3) changing the contact location on the object, and 4) achieving generalized performance across various objects.
To address this problem, we propose a hierarchical reinforcement learning framework.

This framework possesses high-level policy and low-level skills.
The high-level policy decides which skill to choose and the low-level skills generate actual robot actions.
We consider three types of robot actions to deal with the occluded grasping problem, pivoting, pushing, and grasping.
Each low-level skill corresponds to these robot actions.
Since pivoting involves complex physical interactions, we apply domain randomization during training to improve robustness and generalizability.
We also employ Conditional Variational Autoencoder (CVAE) to infer the contact location to successfully execute actions from the low-level skills.
We would like to highlight that our framework does not require any human demonstrations.

Finally, we verify the effectiveness of our framework by conducting numerical simulations and physical experiments.
The results of numerical simulations indicate that our framework achieves the highest success rate of task completion compared to other baselines.
For physical experiments, we aim to perform zero-shot sim-to-real transfer of the trained high-level and low-level policies and demonstrate the generalizability of the proposed framework to unseen objects.

\begin{figure*}[t]
 \centering
    \includegraphics[clip, width=2.00\columnwidth]{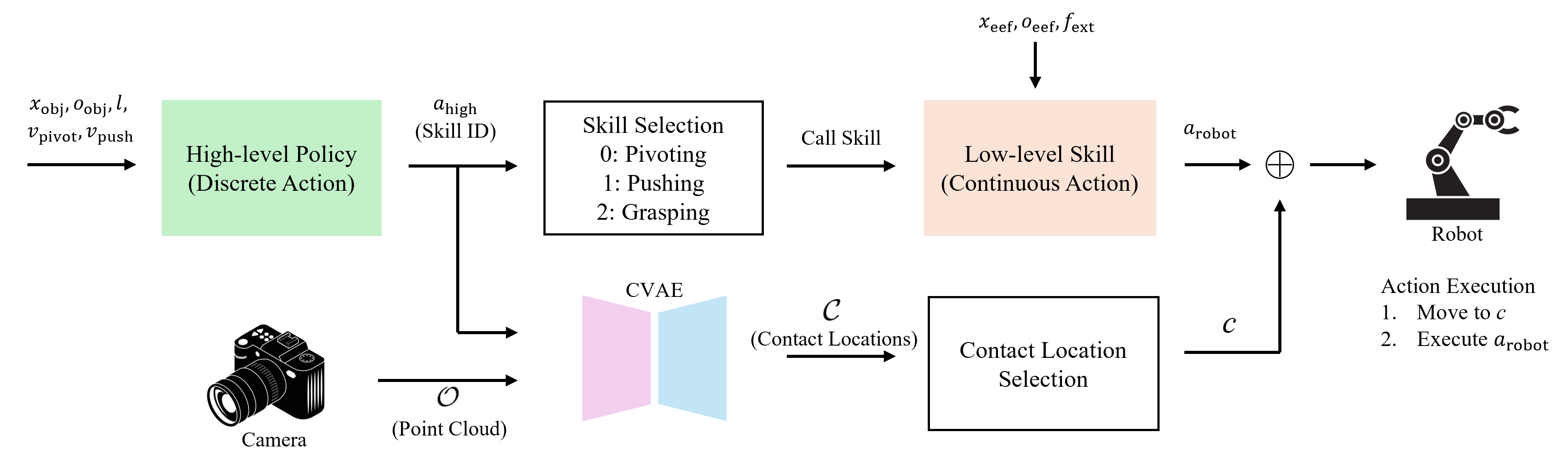}
 \caption{The overview of our hierarchical reinforcement learning framework for occluded grasping tasks. The high-level policy decides which low-level skill to use based on the object pose and subtask completions. Then, the selected low-level skill generates a robot action. To guide the robot to an appropriate contact location for a successful execution of the low-level skill, we adopt CVAE. The contact locations are collected through successful rollouts of the low-level skills and we do not need human demonstrations. The CVAE is conditioned on the skill ID and object point clouds, so that the CVAE can infer the contact location according to the object and the skill. The robot moves to the contact location inferred by the CVAE first, then the robot executes the action from the low-level skill.}
 \label{fig: framework_overview}
\end{figure*}

\section{RELATED WORK}

\subsection{Occluded Grasping with Extrinsic Dexterity}
Occluded grasping problems with a parallel gripper are often solved using extrinsic dexterity \cite{dafle2014extrinsic}, which utilizes extrinsic contacts to enhance the dexterity for robotic manipulation.
The approaches to tackle occluded grasping problems are roughly categorized into the following two: 1) Pivot then grasp, and 2) Tilt then grasp.

The approach of 1) first pivots an object and then grasps the object.
\cite{cheng2022contact, taylor2023object, zhang2023stocs} discuss pivoting problems with model-based approach.
Since the contact dynamics become hybrid due to the different contact modes, model-based approaches often describe such hybrid dynamics as complementarity constraints.
To formulate an accurate model, prior works often impose assumptions when contacts occur, and this limits the applicability of model-based frameworks.
Besides, model-based approaches require physical parameter identifications of objects.

In contrast to the model-based approach, model-free approaches can leave the complex physical dynamics as black-box and directly obtain a control law (policy) through the learning process.
Particularly, model-free reinforcement learning is utilized to handle pivoting and occluded grasping problems. 
\cite{zhang2023learning} proposes a generalizable pivoting framework for boxes, circular shapes, and bottles, but does not consider grasping. 
\cite{zhou2023learning} proposes a framework of occluded grasping problems with pivoting motion for a single robotic manipulator.
In this work, the authors do not explicitly design the robot motion, and the motion is an emergent behavior from the reward.
Since this work assumes grasp accessible walls, the success rate of this framework will be lower on grasp constraining walls.
\cite{yang2024learning, wu2024one} utilize action primitives (push, pivot, grasp) to tackle an occluded grasping problem.
However, \cite{yang2024learning} considers grasp accessible walls and demonstrate the effectiveness of this framework only on box-like objects.
The framework in \cite{wu2024one} relies on human demonstrations to coordinate predesigned action primitives and to select contact locations. 
In contrast to this, our approach does not utilize any human demonstrations, and the framework automatically chooses an appropriate skill and a contact location to complete subtasks.

The approach of 2) considers tilting an ungraspable object against a wall by leveraging extrinsic contacts between the object and the wall, then the object is detached from the surface and becomes graspable \cite{sun2020learning, yamada2025combo, wang2024multi}.
In \cite{sun2020learning}, the authors consider a dual robot framework.
One robotic arm tilts an object against a wall and the other robotic arm grasps the tilted object.
This framework considers occluded grasping problems on an grasp constraining wall, however, it requires two robotic arms.
\cite{wang2024multi} addresses occluded grasping problems on an grasp constraining wall with a single robotic arm.
This approach first tilts an object to a specific pose and changes the gripper's orientation to grasp the object.
When it changes the gripper's orientation, the gripper needs to move on the object while maintaining contact with the object.
This involves complex physical interactions, thus, the success rate of occluded grasping tasks with this framework is relatively low, even for box-like objects.
Besides, this framework requires a manual condition to switch between the tilting and the grasping actions.
In contrast to these previous works, our framework achieves high success rates and demonstrates its generalizability by automatically switching between different types of actions in grasp constraining environments.

\subsection{Contact Location Estimation for Robotic Manipulation}
Generative models such as variational autoencoder (VAE) are widely used as contact location estimators for grasping or manipulation tasks.
\cite{mousavian20196} utilizes VAE to infer grasp candidates of diverse objects.
\cite{pmlr-v205-wu23b, zhang2024multi, zhao2024decomposed} use CVAE and vector-quantized VAE to infer grasping configurations for multi-finger hands.
Our framework also adopts a CVAE, however, our method infers contact locations for multiple subtasks (pivoting, pushing, or grasping), while other frameworks are designed for a single, specific task.
Another approach is to use an object point cloud for contact location selection.
\cite{zhou2023hacman, jiang2024hacman++} propose a non-prehensile manipulation framework by choosing a contact location from the points in a point cloud.
However, these works do not consider manipulation with extrinsic contacts.
In addition to VAE and its variants, utilization of human demonstrations is also a feasible choice.
\cite{liu2024one} adopts human demonstrations to estimate contact locations for different tasks. 
In our framework, the training data is collected through successful rollouts of low-level skills, hence, we do not need human demonstrations.

\section{PROBLEM FORMULATION}
In this section, we address the problem statement that we focus on in this work and introduce the fundamentals of reinforcement learning.

\subsection{Problem Statement}
In this work, we mainly focus on an occluded grasping problem on grasp constraining walls and a hierarchical reinforcement framework to decompose such a long-horizon non-prehensile manipulation problem into simple subproblems to easily handle the original problem.  
We handle an object on a flat surface like a table that is ungraspable at the initial state and a fixed proximity wall perpendicular to the surface.
We assume the lateral direction of the wall is aligned with $y$-axis in the Cartesian coordinate.
The grasp constraining walls are tall, i.e., the object does not become graspable even after the robot pivots the object.

To complete the manipulation task, we aim to learn a policy to manipulate the pose of the object by leveraging extrinsic contact between the wall and the object to make it graspable.
The size and position of the wall vary, and our policy automatically manipulates the object by adapting to the different walls.
We also aim to consider generalizing the policy to various objects with zero-shot sim-to-real transfer.

\subsection{Fundamentals of Reinforcement Learning}
An environment for reinforcement learning is described as a Markov Decision Process (MDP).
MDP is a sequential stochastic process and can be modeled as a tuple $(\mathcal{S}, \mathcal{A}, \mathcal{P}, r_t)$ where $\mathcal{S}$ denotes the state space, $\mathcal{A}$ denotes the action space, $\mathcal{P}(s_{t+1}|s_t, a_t)$ denotes the transition probability, which describes the probability from the current state $s_t$ to the next state $s_{t+1}$ with the action $a_t$, and $r_t$ denotes the reward.
The primary objective of reinforcement learning is to find a policy $\pi(a_t|s_t)$ to maximize the expected return $\mathbb{E}[\sum^{\infty}_{t=0}\gamma^t r_t]$ where $\gamma$ is a discount factor.

\section{METHODOLOGY}
\begin{figure}[t]
 \centering
    \includegraphics[clip, width=0.9\columnwidth]{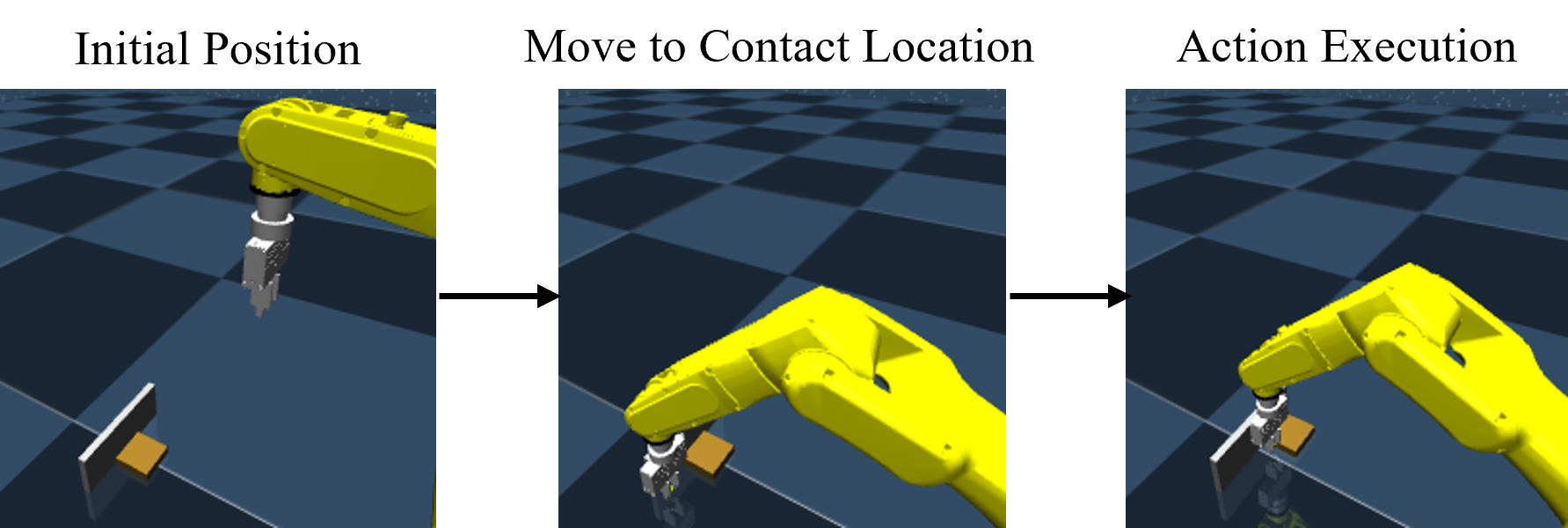}
 \caption{An example for a step of executing the proposed framework. The robot first stays away from the object to get a point cloud, then reaches the contact location. Finally, the robot executes the action.}
 \label{fig: example_step}
\end{figure}

In this section, we introduce the proposed framework to deal with this occluded grasping problem.
The hierarchical reinforcement learning framework comprises a high-level policy to decide which skill to use based on the observation and a selected low-level skill to generate an action for the robot.
Before executing robot actions from the low-level skills, we guide the robot to a desired endeffector pose corresponding to the skills by following a linearly interpolated trajectory.
To infer the desired endeffector pose for pivoting and pushing, we adopt a CVAE conditioned on object point clouds and the skill ID.
Since our main focus is not inferring grasp candidates, we calculate the desired grasping pose by using object geometries.
However, our framework still accepts the existing grasp proposal networks such as \cite{mousavian20196}.
In this work, we use Euler angles as orientations of both objects and the robot.
Note that we train low-level skills first, then CVAE, and finally the high-level policy.
When we train the high-level policy, the low-level skills and CVAE are frozen.
Figure \ref{fig: framework_overview} summarizes the overall proposed framework.
As illustrated in Fig. \ref{fig: example_step}, the robot first stays away from the object to get a point cloud, then moves to the contact location and executes the action sampled from the selected low-level skill.
We further discuss the details of the framework in the following subsections.

\subsection{High-level Policy}
The role of the high-level policy is to choose an appropriate skill based on the observation, thus, the action space is discrete.
Hence, we employ Deep Q Network (DQN) \cite{mnih2013playing} to train the high-level policy.
The action $a_{\mathrm{high}} \in \{0, 1, 2\}$ corresponds to the low-level skill ID where 0 is pivoting, 1 is pushing, and 2 is grasping. 
The observations are the position and orientation of the object $x_{\mathrm{obj}} \in \mathbb{R}^3$, $o_{\mathrm{obj}} \in \mathbb{R}^3$, the lateral length of the wall $l \in \mathbb{R}$, and subtask (pivoting and pushing) completion flags $v_{\mathrm{pivot}} \in \{0,1\}$, $v_{\mathrm{push}} \in \{0,1\}$.
For pivoting, the task completion criterion is $d \leq \frac{10\pi}{180}$ where $d \in \mathbb{R}$ denotes the distance between the current object rotation matrix $R \in \mathbb{R}^{3\times3}$ and the rotation matrix when the object orientation is perpendicular to the table $R_{\mathrm{perp}} \in \mathbb{R}^{3\times3}$, $d = \cos^{-1}( \frac{1}{2}(\mathrm{tr}(R_{\mathrm{perp}}R^\top)-1))$.
For pushing, if the distance from the current object position $x_{\mathrm{obj}} \in \mathbb{R}^3$ to the goal position $x_{\mathrm{goal}} \in \mathbb{R}^3$ is less than 0.03 [m], we regard this as completion.
We calculate $x_{\mathrm{goal}}$ using $l$ so that the object becomes graspable.
The high-level policy can adapt to different goal positions by taking $l$ as an observation.

We construct the following reward to train the policy
\begin{equation}
    \begin{aligned}
    &r = -\| x_{\mathrm{obj}} - x_{\mathrm{goal}} \|_2 + r_{\mathrm{bonus}} + r_{\mathrm{penalty}} + r_{\mathrm{done}} \\[1ex]
    &r_{\mathrm{bonus}} = 
    \begin{cases}
        0.05 & \text{if } d \leq \frac{10\pi}{180} \\[1ex]
        0    & \text{otherwise}
    \end{cases}\\[1ex]
    &r_{\mathrm{penalty}} = 
    \begin{cases}
        -30 & \hspace{-60mm} \ \ \ \ \  \text{if } a_{\mathrm{high}} = 0 \ \text{and}\ v_{\mathrm{pivot}}=1 \ \text{or} \\
        \hspace{15mm} a_{\mathrm{high}} = 1 \ \text{and} \ v_{\mathrm{push}} = 1  \ \text{or} \\
        \hspace{15mm} a_{\mathrm{high}} = 0 \ \text{and} \ v_{\mathrm{push}} = 0 \ \text{or} \\
        \hspace{15mm} a_{\mathrm{high}} = 2 \ \text{and} \ v_{\mathrm{pivot}}*v_{\mathrm{push}} = 0 \\[1ex]
        0     & \hspace{-60mm} \ \ \ \ \ \text{otherwise}
    \end{cases}\\[1ex]
    &r_{\mathrm{done}} = 
    \begin{cases}
        1.05 & \text{if } \text{grasping} \\[1ex]
        0     & \text{otherwise}
    \end{cases}
    \end{aligned}
    \label{eq: reward_high_level}
\end{equation}
When $d = 0$, the object stands on the table, hence the robot completes the pivoting task. 
We assign a buffer of 10 [deg] for a pivoting success criterion and use this criterion to feed the bonus $r_{\mathrm{bonus}}$ and penalty $r_{\mathrm{penalty}}$ to the high-level agent.
Since the dense reward is calculated by the distance from the current object position to the goal object position, pivoting success itself does not increase the reward.
To encourage the robot to pivot an object, we feed the bonus reward of 0.05 when the robot succeeds in pivoting an object.
We also feed a penalty reward of -30 as described in Eq. \ref{eq: reward_high_level}.
The penalty is imposed when the high-level policy chooses a skill that corresponds to the completed subtask or chooses the grasping skill without completing other subtasks.
Although the actual values of the bonus and the penalty need not be fixed to these specific values, we indicate that adding such bonus and penalty terms accelerates the training procedure of the high-level policy.
We feed the done reward of 1.05 to the agent when it succeeds in grasping the object.

\subsection{Low-level Skill}
The lower-level skills generate an actual robot action $a_{\mathrm{robot}}$.
In this work, we consider the following three skills: pivoting, pushing, and grasping.

For the pivoting skill, we handle a continuous action space to complete a complex contact-rich manipulation task.  
We employ Soft Actor-Critic (SAC) \cite{haarnoja2018soft} and design the following reward to train the pivoting skill.
\begin{align}
    r_{\mathrm{pivot}} = \frac{\pi}{2} - d
\end{align}
The observations for this skill are the position and orientation of the endeffector $x_{\mathrm{eef}} \in \mathbb{R}^3$, $o_{\mathrm{eef}} \in \mathbb{R}^3$, and the external contact force $f_{\mathrm{ext}} \in \mathbb{R}^6$.
The pivoting skill generates an action of an endeffector velocity in the Cartesian space and we do not consider any rotation of the endeffector during the pivoting task.
Thus, the action from the pivoting skill $a_{\mathrm{pivot}} \in \mathbb{R}^3$ should be in a three dimensional space.
To acquire the generalizability, we apply domain randomization.
We add a zero-mean Gaussian noise $\mathcal{N}(0,0.2I)$ to $f_{\mathrm{ext}}$, $x_{\mathrm{eef}}$, and $o_{\mathrm{eef}}$.

For pushing and grasping, we utilize hand-crafted skills due to the simplicity of tasks.
The pushing skill feeds a constant action of the endeffector velocity $a_{\mathrm{push}} = [0, -0.005, 0]$ to the robot and the robot pushes the object after moving to the contact location inferred by CVAE.
The grasping skill feeds a constant action of the endeffector velocity $a_{\mathrm{grasp}} = [0, 0, -0.01]$.
When the robot grasps the object, the robot first moves above the grasp location, then gradually approaches to the object to grasp with $a_{\mathrm{grasp}}$.

\subsection{Conditional Variational Autoencoder}
To estimate the desired contact location for execution of each skill, we adopt CVAE.
We feed object point clouds $\mathcal{O} \in \mathbb{R}^{n\times m}$ and skill IDs $a_{\mathrm{high}}$ as conditions, and feed desired contact locations $\mathcal{C} \in \mathbb{R}^{n\times3}$ as data to train CVAE where $n$ denotes the number of object point clouds and desired contact locations, and $m$ denotes the length of a flattened array of a point cloud data.
The desired contact locations are collected by the successful rollouts of the skills.
When we run CVAE online, the point cloud is provided by a depth camera and the skill ID is provided by the high-level agent.
When we train CVAE, we add a zero-mean Gaussian noise $\mathcal{N}(0, 0.003I)$ to the object point cloud to simulate the noise and acquire robustness.

In the training process of CVAE, CVAE reconstructs the probability distribution of the original data $p(\mathcal{C}|\mathcal{O}, a_{\mathrm{high}})$.
Mathematically, $p(\mathcal{C}|\mathcal{O}, a_{\mathrm{high}})$ is calculated through 
\begin{align}
    p(\mathcal{C}|\mathcal{O}, a_{\mathrm{high}}) = \int p(\mathcal{C}|z, \mathcal{O}, a_{\mathrm{high}}) p(z|\mathcal{O}, a_{\mathrm{high}}) dz
\end{align}
However, this integral is computationally intractable.
Thus, CVAE possesses an encoder and a decoder network to approximately reconstruct the distribution $p(\mathcal{C}|\mathcal{O}, a_{\mathrm{high}})$ by maximizing the Evidence Lower Bound $L$.
\begin{equation}
    \begin{aligned}
      L =  & - D_{KL}(q(z|\mathcal{C}, \mathcal{O}, a_{\mathrm{high}}) \| p(z| \mathcal{O}, a_{\mathrm{high}})) \\
         & + \mathbb{E}_{z\sim q(z|\mathcal{C}, \mathcal{O}, a_{\mathrm{high}})}[\log(p(\mathcal{C}|z, \mathcal{O}, a_{\mathrm{high}}))] 
    \end{aligned}
\end{equation}
The encoder $q(z|\mathcal{C}, \mathcal{O}, a_{\mathrm{high}})$ maps contact locations $\mathcal{C}$ to the latent space, and the latent variable $z$ is subject to a standard Gaussian distribution $\mathcal{N}(0,I)$.
The decoder $p(\mathcal{C}|z, \mathcal{O}, a_{\mathrm{high}})$ reconstructs $\mathcal{C}$ with a given $z$ and conditions.

We pick up a contact location $c \in \mathbb{R}^3$ to guide the robot to a specific location.
We empirically find that the lowest $z$ position is effective for pivoting action, and the largest $y$ position is effective for pushing action.
Hence, we choose a pivoting contact location that has the lowest $z$ position among the inferred contact locations, and we choose a pushing contact location that has the largest $y$ position among the inferred contact locations.
When we run the CVAE while executing the whole pipeline, we add an offset vector to the inferred contact location.

\section{EXPERIMENTS}
In this section, we conduct numerical simulations and physical experiments to verify the effectiveness of the proposed framework.

\subsection{Numerical Simulation}
To begin with, we conduct numerical simulations to verify the effectiveness of our proposed framework.
As mentioned in Section III, we consider an environment where there is an object and a fixed wall, and the fixed wall is grasp constraining.
For the manipulated object, we consider different size of boxes.
To evaluate the performance, we adopt the following baseline methods for comparisons.
\begin{itemize}
    \item SAC \cite{haarnoja2018soft}: This baseline is designed for pivoting tasks and adopted from \cite{zhang2023learning, zhang2023efficient}.  
    \item Proposed w/o CVAE: This approach uses the same architecture of the proposed method without CVAE.
    \item Proposed w/o skills: This approach uses a SAC-based framework to obtain a continuous action. The robot randomly picks up a contact location from the outputs of the CVAE and the grasping location. 
    \item Point Cloud: This method uses an object point cloud to estimate a contact location instead of CVAE. We adopt DQN and choose a contact location from the points in a point cloud that correspond to the highest Q value. For a fair comparison, we adopt the same skills (pivoting, pushing, and grasping). The observation is an object point cloud and the vector from the current object pose to the goal object pose. This baseline is inspired by HACMAN \cite{zhou2023hacman} and HACMAN++ \cite{jiang2024hacman++}.
    \item Ungraspable \cite{zhou2023learning}: This approach leverages extrinsic dexterity to grasp an ungraspable object. Utilizing extrinsic dexterity is an emergent behavior of the robot (not explicitly designed) and the robot tilts or pivots the object to make it graspable. This work implicitly assumes a grasp accessible environment for manipulation, thus we aim to verify the performance in an grasp constraining environment. We use the provided implementation from the authors of \cite{zhou2023learning} to get the results. 
\end{itemize}
\begin{figure}[t]
 \centering
    \includegraphics[clip, width=0.99\columnwidth]{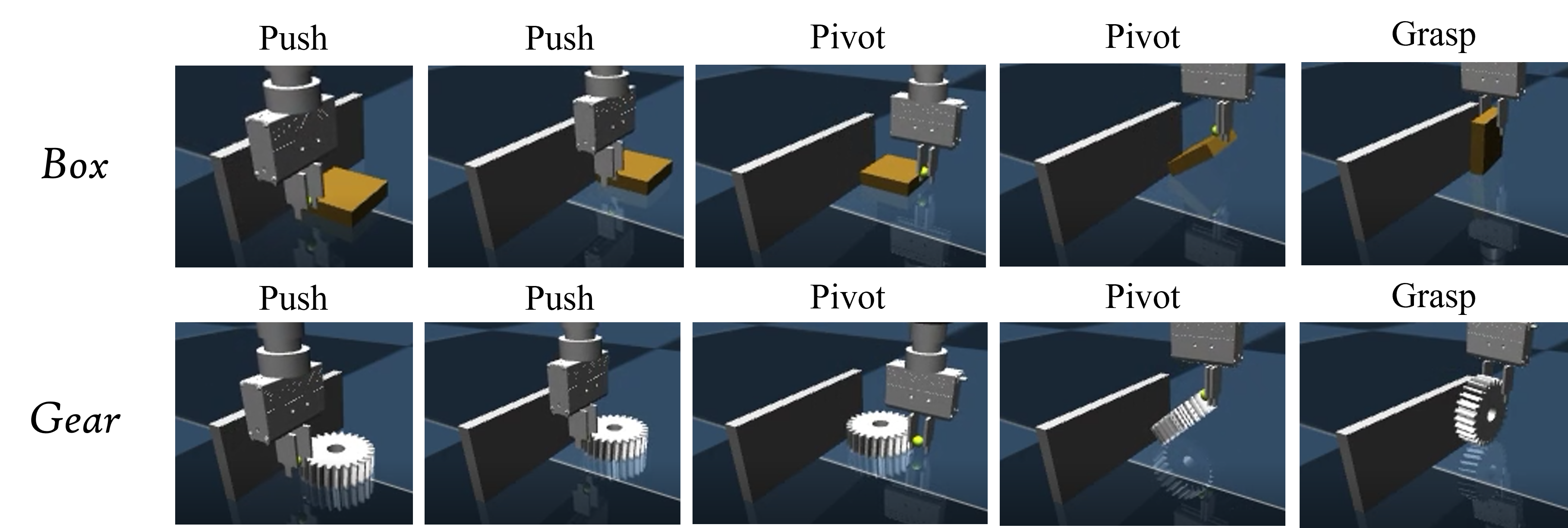}
 \caption{An example of a successful action sequence of the robot in MuJoCo simulation for a box and a gear object. The proposed framework can manipulate the objects to graspable poses by combining pivoting, pushing, and grasping actions.}
 \label{fig: simulation_env}
\end{figure}
\begin{figure}[t]
 \centering
    \includegraphics[clip, width=0.9\columnwidth]{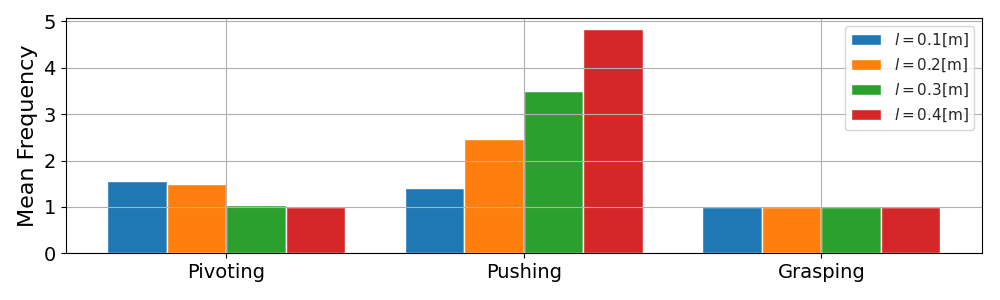}
    \caption{A mean frequency of skill selection for different lengths of the grasp constraining walls over 30 episodes. }
    \label{fig: skill_frequency}
\end{figure}
\begin{table}[b]
  \centering
  \begin{threeparttable}[b]
    \caption{Success rates on grasp accessible and grasp constraining walls}
    \vspace{-0.1in}

    \begin{tabular*}{64mm}{c|cc}
      \Xhline{1pt}
                    & Accessible &  Constraining \\ \hline
      SAC                  &  100\% &    0\% \\
      Proposed w/o CVAE    & 0\% &   0\% \\
      Proposed w/o skills  & 0\% &   0\% \\
      Point Cloud               &   13.3\% &  13.3\% \\
      Ungraspable          & 100\% & 0\% \\
      Proposed                 & 100\% & 100\% \\
      \Xhline{1pt}
    \end{tabular*}
    \label{tab: result_sim}
    \vspace{-0.1in}
  \end{threeparttable}
\end{table}
We train both the high-level policy and the low-level pivoting skill using the implementations from RLkit with a batch size of 256 for the high-level agent and 4096 for the low-level pivoting skill.
The CVAE consists of an encoder with two-layer ReLU networks with 512 units, a dropout layer, and a decoder with the same architecture as the encoder.
The batch size is 256 and the learning rate is $10^{-5}$.
We use 30,000 contact locations and object point clouds for pivoting and pushing to train the CVAE, and run 1,000,000 steps.

For simulation settings, we use a $0.1 \times 0.1 \times 0.03 \ [\mathrm{m}^3]$ box for a manipulated object and vary the size of object to examine the generalizability of the proposed framework and baselines.
The range is $\mathrm{size}_x \in [0.08 \ [\mathrm{m}],0.12 \ [\mathrm{m}]]$, $\mathrm{size}_y \in [0.08 \ [\mathrm{m}],0.12 \ [\mathrm{m}]]$.
In this work, we only consider task success rate as an evaluation metric.
We allow the high-level policy to choose the skills within 15 trials and if the robot completes the occluded grasping task, we regard this as success.
We run the proposed framework and the baselines 30 times. 
Figure \ref{fig: simulation_env} depicts the successful action sequence in the simulation environment.
We execute the proposed framework and the baselines on grasp accessible and grasp constraining walls to demonstrate the effectiveness of the proposed method.

Table. \ref{tab: result_sim} exhibits the simulation results.
In this work, we focus on the task success rate and do not consider any other metrics.
Note that for grasp accessible walls, we regard it as success if the robot pivots the object because the pose becomes graspable.
We observe that some of the baselines demonstrate high success rates on grasp accessible walls, however, the success rates become lower without a CVAE or skills.
Without a CVAE, it is hard to find an appropriate contact location to execute an action from a skill such as pushing and pivoting.
Besides, without skills, it is difficult to learn actions to appropriately manipulate an object, particularly when the task requires extrinsic contacts.
On the other hand, on grasp constraining walls, the performance of all baselines degrades significantly.
The reason is that without skills or contact location estimators, finding an appropriate contact location and an effective action to complete the task is difficult.
While the Point Cloud baseline has both skills and a contact location selection algorithm, we observe that simultaneously learning contact locations and skill coordination during training is difficult.
When the robot executes the Ungraspable framework, the robot tends to grasp the object after pivoting the object, though the pivoting motion is emergent behavior and we do not explicitly design the pivoting motion of the robot.
Hence, when the wall becomes constraing, the robot fails to grasp the object.
These results demonstrate the effectiveness of our framework, particularly in grasp constraining environments.

Then, we conduct detailed analyses of the performance of the proposed framework.
In these analyses, we apply the proposed framework for unseen boxes and a circular object to verify the generalizability.
We use a large box ($0.13 \times 0.13 \times 0.036 \ [\mathrm{m}^3]$), a small box ($0.072 \times 0.072 \times 0.036 \ [\mathrm{m}^3]$), a long box ($0.12 \times 0.06 \times 0.036 \ [\mathrm{m}^3]$), a short box ($0.06 \times 0.12 \times 0.036 \ [\mathrm{m^3}]$), and a gear (a circular shape object) ($0.06^2 \pi \times 0.026 \ [\mathrm{m}^3]$).
Besides, we change the lateral length of the wall $l$ from 0.10 $[\mathrm{m}]$ to 0.40 $[\mathrm{m}]$.
To evaluate the success rate for each simulation setting, we execute the framework 30 times.
We allow the high-level policy to choose the skills within 15 trials, and if the robot completes the occluded grasping task, we regard this as a success.
We summarize the results of these analyses in Table \ref{tab: result_generalizability}.
In most cases, our framework achieves 100 \% success rates, and this demonstrates the robustness to unseen objects and environmental differences.
We also analyze the frequency of skill selection and summarize the results in Fig. \ref{fig: skill_frequency}.
We recognize that the frequency of the pushing skill becomes higher as the length of the wall becomes longer.
This demonstrates the generalizability of the high-level policy to different environments.
Besides, we observe that the frequency of the pivoting becomes larger than 1.
When the robot fails to pivot an object, the high-level policy chooses the pivoting skill again to complete the task.
Hence, we can interpret this as a recovery behavior.

\begin{table}[b]
  \centering
  \begin{threeparttable}[b]
    \caption{Success rate for unseen objects and different walls of the proposed framework}
    \vspace{-0.1in}

    \begin{tabular*}{83mm}{c|cccc}
      \Xhline{1pt}
                 & $l=0.1$[m] & $l=0.2$[m] & $l=0.3$[m] & $l=0.4$[m] \\ \hline
        Large Box  & 96.7 \% & 100 \% & 100 \% & 93.3 \% \\
        Small Box  & 100 \% & 100 \% & 100 \% & 100 \% \\
        Long Box  & 100 \% & 100 \% & 100 \% & 100 \% \\
        Short Box & 100 \% & 100 \% & 100 \% & 100 \% \\
        Gear & 90 \% & 100 \% & 100 \% & 100 \%   \\
      \Xhline{1pt}
    \end{tabular*}
    \label{tab: result_generalizability}
  \end{threeparttable}
\end{table}

\subsection{Physical Experiments}
\begin{table}[b]
  \centering
  \begin{threeparttable}[b]
    \caption{Results of real-world physical experiments}
    \vspace{-0.1in}

    \begin{tabular*}{67mm}{c|cc}
      \Xhline{1pt}
                 & Size $[\mathrm{m}^3]$ & Success \\ \hline
        Box (Large)  & $0.11 \times 0.142 \times 0.038$ & 10/10 \\
        Box (Medium)  & $0.1 \times 0.101 \times 0.04$  & 10/10 \\
        Box (Small) & $0.064 \times 0.121 \times 0.027$  & 8/10 \\
        Bottle (Large) & $0.068 \times 0.212 \times 0.04$ & 10/10 \\
        Bottle (Small) & $0.055 \times 0.159 \times 0.022$  & 7/10 \\
        Circle & $0.0545^2 \pi \times 0.026$  & 9/10 \\
        Average & N/A & 9/10 \\
      \Xhline{1pt}
    \end{tabular*}
    \label{tab: result_experiments}
    \vspace{-0.1in}
  \end{threeparttable}
\end{table}
\begin{figure}[t]
 \centering
    \includegraphics[clip, width=0.99\columnwidth]{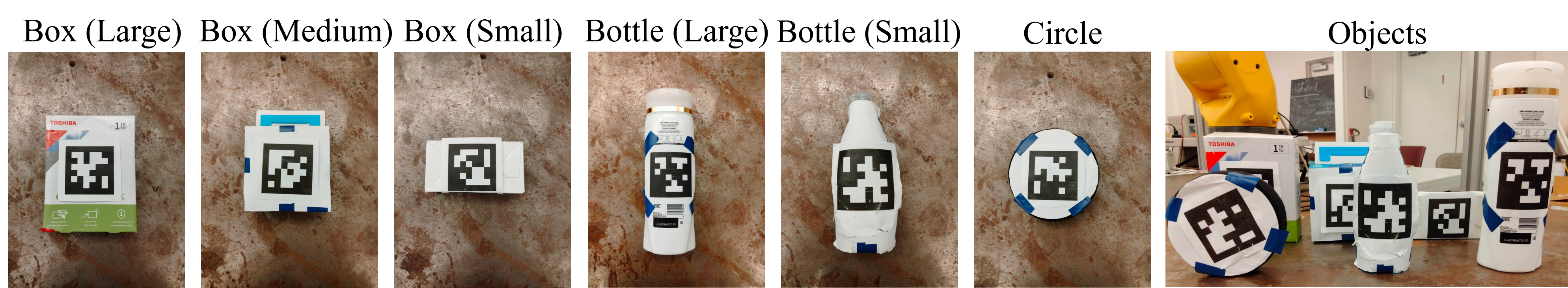}
 \caption{Test Objects for Real-world Experiments. All objects are out-of-distribution. Particularly for bottle objects and a circular object, their curved surfaces make it difficult for a robot to maintain contacts during manipulation.}
 \label{fig: objects_realworld}
\end{figure}
\begin{figure}[t]
 \centering
    \includegraphics[clip, width=0.99\columnwidth]{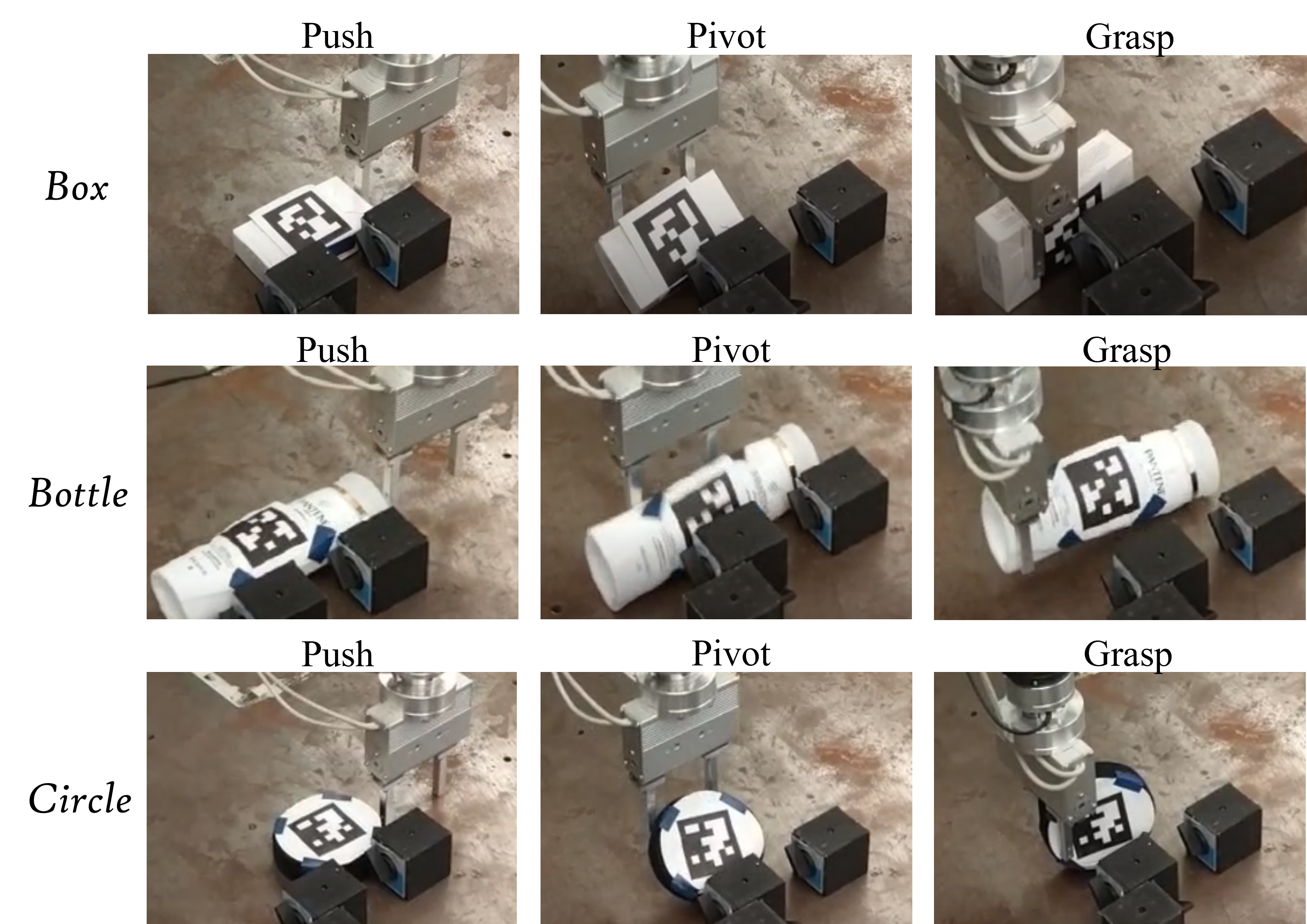}
 \caption{Examples of manipulation scenarios of physical experiments. The proposed framework realizes sim-to-real transfer in pivoting, pushing, and grasping for various objects.}
 \label{fig: physical_experiment}
\end{figure}

We also conduct physical experiments in addition to numerical simulations.
In these experiments, we aim for sim-to-real zero-shot transfer of the trained policy.
We use an RGB-D camera (Ensenso N35) to obtain an object point cloud and apply a color-based filter to segment the point cloud.
We also use AprilTag \cite{olson2011apriltag} to estimate the object pose.
We adopt a 6-DoF robotic manipulator with a two-finger parallel gripper that makes it difficult to grasp a large flat object without manipulating its pose.
We control the robot using a Cartesian space admittance control, and the contact force on the endeffector is captured by a force/torque sensor mounted on the wrist of the endeffector.
To compensate for the difference between a low-level controller in simulation and the real world, we uniformly rescale the action from the low-level skills. 
We use two magnet bases to make a wall. 
Due to the position of the camera, vision occlusion problems occur if we make a tall wall to cover large objects.
However, both on a tall wall and a short wall, the actual physical interaction happens on the lower side of the wall.
Hence, the performance of the trained RL policy remains unchanged and we can still verify the performance of the proposed framework in real world.

We handle six different objects (3 boxes, 2 bottles, and 1 circular object) to assess the generalization performance of the proposed framework.
Note that the boxes are out-of-distribution because the sizes are not in the range of those used in the simulation.
Other objects are unseen since we do not use them for training.
For bottle objects and the circular object, the curved surfaces make it difficult for the robot to maintain contact during manipulation.

The successful examples are visualized in Fig. \ref{fig: physical_experiment}.
The robot successfully utilizes extrinsic contacts to manipulate the pose of the box, the bottle, and the circle, and grasp them.
All experimental results are summarized in Table. \ref{tab: result_experiments}.
As shown in this table, our approach achieves average success rates of 90 \% in real-world experiments, demonstrating the generalizability of the proposed framework.
Particularly, while occluded grasping of bottles and a circular object is very difficult, our framework still achieves high success rates.

\section{CONCLUSION}
In this study, we present a hierarchical reinforcement learning framework for occluded grasping problems on grasp constraining walls.
The proposed framework performs well particularly in grasp constraining environments compared with other baselines even though the task comprises complex physical interactions between the environment and the robot.
We conducted physical experiments to demonstrate the zero-shot sim-to-real performance of the proposed framework, and the framework achieved an average success rate of 90\% in the occluded grasping task for six out-of-distribution objects.

As future work, we will tackle more complex problems such as occluded grasping in a bin with grasp constraining walls.

\begingroup
\bibliographystyle{IEEEtran}
{\small\bibliography{Reference.bib}}
\endgroup
\vspace{12pt}

\end{document}